\documentclass[conference]{IEEEtran}
\IEEEoverridecommandlockouts
\usepackage{cite}
\usepackage{amsmath,amssymb,amsfonts}
\usepackage{algorithmic}
\usepackage{graphicx}
\usepackage{textcomp}
\usepackage{xcolor}
\usepackage{booktabs}
\usepackage{multirow}
\usepackage[table]{xcolor}
\usepackage{float}
\usepackage{stfloats}
\definecolor{lightgreen}{RGB}{220, 240, 220}
\definecolor{lightyellow}{RGB}{255, 250, 220}
\def\BibTeX{{\rm B\kern-.05em{\sc i\kern-.025em b}\kern-.08em
    T\kern-.1667em\lower.7ex\hbox{E}\kern-.125emX}}
\begin{document}

\title{Hand3R: Online 4D Hand-Scene Reconstruction in the Wild}

\author{Wendi Hu, Haonan Zhou, Wenhao Hu, Gaoang Wang \\ Zhejiang University}

\maketitle

\begin{abstract}
For Embodied AI, jointly reconstructing dynamic hands and the dense scene context is crucial for understanding physical interaction. However, most existing methods recover isolated hands in local coordinates, overlooking the surrounding 3D environment. To address this, we present Hand3R, the first online framework for joint 4D hand-scene reconstruction from monocular video. Hand3R synergizes a pre-trained hand expert with a 4D scene foundation model via a scene-aware visual prompting mechanism. By injecting high-fidelity hand priors into a persistent scene memory, our approach enables simultaneous reconstruction of accurate hand meshes and dense metric-scale scene geometry in a single forward pass. Experiments demonstrate that Hand3R bypasses the reliance on offline optimization and delivers competitive performance in both local hand reconstruction and global positioning.
\end{abstract}


\section{Introduction}
\label{sec:intro}

Hands serve as the primary interface through which humans perceive, manipulate, and reshape the physical world. Consequently, understanding the complex dynamic interplay between hands and 3D scenes is fundamental to a wide range of applications, including AR/VR, Embodied Intelligence, and Human-Computer Interaction. To fully unlock these capabilities, the ultimate goal of computer vision should extend beyond merely estimating the isolated hand pose and shape; it necessitates a holistic understanding that simultaneously recovers the hand's exact metric location within the global world frame and the dense geometry of its surrounding environment.

\begin{figure}[t]
    \centering
    \includegraphics[trim={150pt 260pt 50pt 180pt}, clip, width=1.2\linewidth]{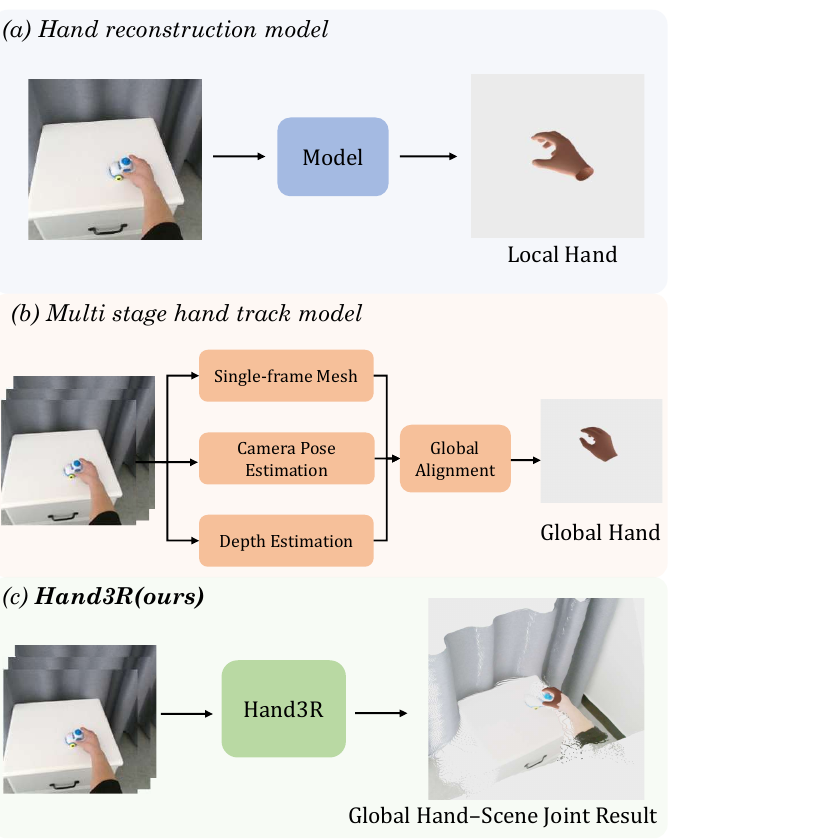} 
    \caption{\textbf{Hand3R vs Other hand methods.} (a): Traditional methods recover isolated hand meshes in root-relative coordinates. (b): Recent global methods rely on complex, disjoint pipelines to lift hands into world space.(c) Hand3R: The first online, end-to-end framework that performs joint 4D hand-scene reconstruction. Hand3R simultaneously recovers accurate global hand trajectories and dense metric-scale scene geometry in a single forward pass.}
  \label{fig:teaser}
    \label{fig:vs}
\end{figure}

As shown in Fig.~\ref{fig:vs}, despite significant strides in monocular 3D hand mesh recovery~\cite{pavlakos2024reconstructing}\cite{potamias2024wilor}, existing approaches predominantly operate in a root-relative camera coordinate system.
While they excel at recovering local articulation, they fundamentally overlook the hand's absolute spatial placement within the global 3D world.
Although recent works like HaWoR\cite{zhang2025hawor} have attempted to recover global hand positioning, they rely on fragmented multi-stage pipelines and do not simultaneously reconstruct the dense 3D scene geometry.
Without the scene context, the reconstructed hands remain spatially isolated, limiting the system's ability to reason about physical interactions, collisions, or the hand-scene relationship essential for embodied tasks.

On one hand, State-of-the-Art (SOTA) local hand experts have mastered fine-grained articulation but operate in a vacuum, lacking world-space awareness. On the other hand, emerging global scene learners have mastered metric-scale mapping but lack the feature resolution to capture the sub-centimeter dexterity of hands.
Therefore, the core research challenge is finding an efficient mechanism to bridge these two distinct domains: How can we seamlessly fuse the high-fidelity priors of a pre-trained hand expert into the consistent metric space of a 4D scene learner in a unified, end-to-end framework?

To address the aforementioned issues, we present \textbf{Hand3R}, the first online framework that synergizes strong local hand priors with global scene context.
Our key insight is that optimal reconstruction is achieved by anchoring the specialized knowledge of a hand expert within the persistent memory of a scene foundation model. 
Instead of treating the two tasks in isolation, we leverage a scene-aware visual prompting mechanism.
Specifically, we extract high-fidelity features from a frozen hand expert and fuse them with spatially-aligned tokens from the scene backbone. We treat these fused features as prompts that are injected into the scene's recurrent state.
This design is highly efficient: it allows Hand3R to inherit the millimeter-level precision of the hand expert and the metric-scale consistency of the scene learner simultaneously. Crucially, this unification is achieved without compromising the reconstruction quality of either domain; instead, it enables robust joint reasoning where both tasks benefit from the shared spatiotemporal context.

In summary, our main contributions are as follows:
\begin{itemize}
    \item \textbf{First Online End-to-End Hand-Scene Reconstruction Framework.} 
    We present Hand3R, the first online, end-to-end framework for joint metric-scale hand and dense scene reconstruction from monocular video.
    
    \item \textbf{Scene-Aware Visual Prompting Mechanism.} 
    We leverage a scene-aware visual prompting mechanism to effectively inject high-fidelity hand priors into the persistent scene memory. This design achieves great hand precision while maintaining the metric consistency of the global environment.
    
    \item \textbf{Competitive Performance.} 
    Experiments on DexYCB and HOI4D datasets demonstrate that Hand3R delivers competitive accuracy in both local hand reconstruction and global positioning.
\end{itemize}

\section{Related-work}
\label{sec:relat}

\subsection{Local Hand Mesh Recovery}
\label{sec:local}
Recovering 3D hand pose and shape from RGB images has been widely studied for over a decade.
In the pioneering work of Boukhayma\cite{boukhayma20193d}, the authors proposed the first fully learnable pipeline to directly regress MANO parameters from RGB images using a CNN backbone.
Several methods have followed this parametric regression paradigm~\cite{zhang2019end}\cite{hasson19_obman}, while others~\cite{kulon2020weakly}\cite{moon2020i2l} introduced non-parametric approaches that directly predict 3D vertices or heatmaps to achieve better image alignment.
Recently, the importance of data and model scaling has been extensively highlighted, with Vision Transformers (ViT) being introduced to enhance reconstruction quality. 
Architectures like MeshTransformer~\cite{lin2021end} and MeshGraphormer~\cite{lin2021mesh} leveraged global attention to model vertex-vertex interactions.
In particular, HaMeR~\cite{pavlakos2024reconstructing} demonstrated that utilizing a pre-trained large-scale ViT and scaling the training data can effectively improve performance in the wild.
This direction has been further refined by WiLoR~\cite{potamias2024wilor}, which adopts Low-Rank Adaptation for efficient hand fine-tuning.
However, despite these improvements in local geometry, these methods operate in a root-relative camera space. They fundamentally lack awareness of the absolute metric scale and the global 3D environment, leaving the reconstructed hands spatially ungrounded.

\subsection{Human-Scene and Hand-Scene Reconstruction}
\label{sec:human}
Recovering dynamic humans within a global 3D scene has been a longstanding goal in computer vision.
Existing methods for joint human-scene reconstruction typically perform global optimization over camera poses, pre-reconstructed scenes, and SMPL mesh parameters inferred from multi-view images~\cite{pavlakos2022theone}\cite{zhao2024synergistic}, often regularized by learned motion priors~\cite{yuan2022glamr}\cite{rempe2021humor}.
Recently, optimization-free approaches have emerged: HAMSt3R~\cite{rojas2025hamst3r}, for example, jointly reconstructs the scene and DensePose from multi-view images in a feed-forward manner.
For monocular video settings, JOSH3R~\cite{liu2025joint} and Human3R~\cite{chen2025human3r} have pushed the boundary towards online processing, eliminating heavy offline dependencies and enabling real-time recovery of global human motion and dense scene geometry in a single forward pass. 

In the domain of global hand analysis, HaWoR~\cite{zhang2025hawor} recently attempted to bridge the gap between local hands and world space by reconstructing hand motion trajectories in global coordinates. 
However, unlike the unified frameworks in body-centric research, HaWoR relies on a complex multi-stage pipeline that decouples the task into separate SLAM-based camera tracking, metric depth estimation, and hand reconstruction. Crucially, it focuses solely on the temporal trajectory of the hands and does not reconstruct the dense 3D scene geometry, limiting the system's ability to reason about physical interactions between the hand and the environment.

In contrast to the rapid progress in body-centric research, Hand-Scene Reconstruction remains an unexplored frontier.
To the best of our knowledge, there exists no unified model capable of jointly recovering metric-scale hand motion and dense scene geometry. Consequently, how to effectively leverage the strengths of mature local hand experts within a global scene reconstruction framework remains an open problem. 
Hand3R addresses this by proposing a unified fusion architecture.

\begin{figure*}[t] 
  \centering
  \includegraphics[width=1.05\linewidth, trim=0cm 4.2cm 0.0cm 3.5cm, clip]{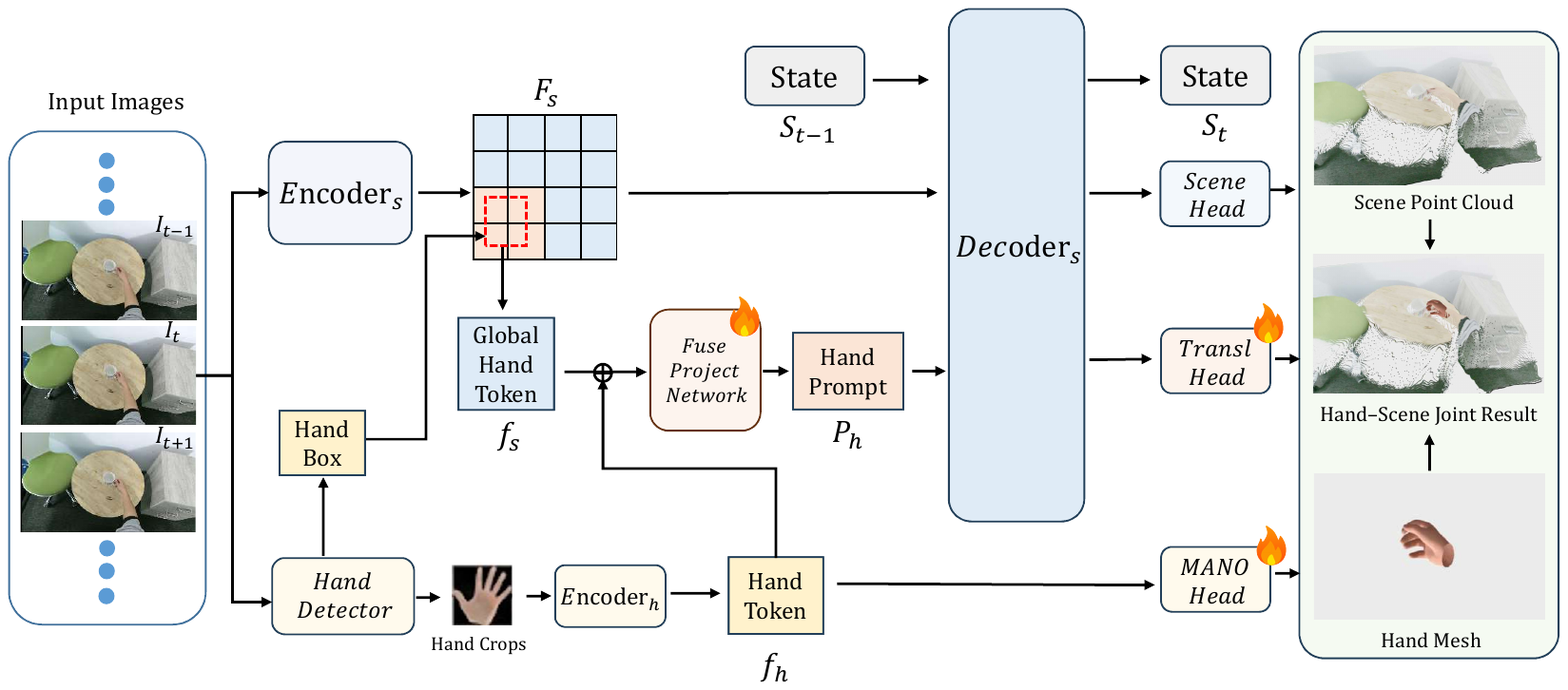}
  
  \caption{\textbf{Overview of the Hand3R framework}. Given a monocular video stream, our model performs online 4D reconstruction in a single forward pass. The pipeline consists of three key stages:
  (1) Dual-stream encoding: We employ a scene encoder to extract dense metric-scale features $\mathbf{F}_s$ and a hand encoder to obtain high-fidelity hand tokens $\mathbf{f}_h$.
  (2) Scene-aware visual prompting: We extract local scene context $\mathbf{f}_s$ by detected hand box and fuse it with the hand token to generate a prompt $\mathbf{P}_h$.
  (3) State-aware decoding and prediction: Prompt is injected into the scene decoder, which interacts with the persistent temporal state $\mathbf{S}_{t-1}$ and dense scene features $\mathbf{F}_{s}$. Then three decoupled heads output the dense scene point cloud, hand position and hand mesh respectively. Finally, we obtain the unified Hand-Scene Joint Result.}
  
  \label{fig:framework}
\end{figure*}

\section{Method}
\label{sec:method}

To capture dynamic hand motions within a globally consistent world frame, we present Hand3R, an online framework that jointly reasons about fine-grained hand geometry and large-scale scene structure. Operated on a continuous RGB stream $\mathbf{I}_t \in \mathbb{R}^{W \times H \times 3}$, our model performs feed-forward inference to simultaneously estimate the camera trajectory $\mathbf{T}_t \in SE(3)$, the dense canonical scene point cloud $\mathbf{X}_t$, and a set of interacting hand meshes $\{\mathbf{M}_t^h\}_{h=1}^H$.

\subsection{Hand3R Architecture}
\label{sec:hand3r}
\subsubsection{\textbf{Overview}}
As illustrated in Fig.~\ref{fig:framework}, Hand3R adopts a dual-stream architecture designed to disentangle the estimation of local hand articulation from global spatial positioning.
Given an input video stream, the system operates in two parallel branches: a scene stream that maintains a persistent 4D metric state, and a hand stream that extracts fine-grained expert features.
These streams converge at the scene-aware prompt fusion module, which injects hand priors into the scene's memory.
Finally, we employ a decoupled prediction strategy: local pose parameters are regressed from the expert branch, while global translation is decoded from the unified scene-hand state.

\subsubsection{\textbf{Dual-Stream Encoding and Scene-Aware Prompt Construction}}
\label{sec:method_prompt}
The core of Hand3R is to bridge the gap between local hand dexterity and global metric awareness. Instead of treating feature extraction and fusion as separate steps, we leverage a unified scene-aware prompt construction mechanism that synergizes signals from two parallel streams.

Specifically, given an input frame $\mathbf{I}_t$, the scene stream utilizes the scene backbone $Encoder_s$ to encode the full image into metric-scale scene tokens:
\begin{equation}
    \mathbf{F}_s = Encoder_s(\mathbf{I}_t) 
\end{equation}

Simultaneously, in the hand stream, for a detected hand with bounding box $\mathbf{b}$, we crop the image region $\mathcal{T}_{crop}(\mathbf{I}_t, \mathbf{b})$ and feed it into a frozen hand expert encoder $Encoder_h$. This yields the high-fidelity hand features:
\begin{equation}
    \mathbf{f}_h = Encoder_h(\mathcal{T}_{crop}(\mathbf{I}_t, \mathbf{b}))
\end{equation}

To ground these floating hand features into the metric world, we extract the local environmental context $\mathbf{f}_{s}$ by performing region-based average pooling on the scene feature map $\mathbf{F}_{s}$ within the bounding box $\mathbf{b}$:
\begin{equation}
    \mathbf{f}_{s} = \mathcal{P}_{avg}(\mathbf{F}_{s}, \mathbf{b})
\end{equation}

Finally, we synthesize the unified hand prompt $\mathbf{p}_h$ by fusing the expert priors with the spatial context:
\begin{equation}
    \mathbf{p}_h = \mathcal{M}_{prompt}(\text{Concat}(\mathbf{f}_{h}, \mathbf{f}_{s}))
\end{equation}

This hybrid prompt $\mathbf{p}_h$ effectively encodes "what the hand looks like" via $\mathbf{f}_{h}$ while being anchored to its specific location in the metric scene via $\mathbf{f}_{s}$, serving as a robust query for the subsequent decoding stage.

\subsubsection{\textbf{State-Aware Decoding and Decoupled Prediction}}

To effectively leverage temporal context while preserving fine-grained articulation details, we design a unified decoding process followed by a decoupled prediction strategy. As illustrated in Eq.~\ref{eq:decoding_state}, the shared decoder takes the generated hand prompts $\mathbf{p}_h$ and interacts with the persistent scene state $\mathbf{S}_{t-1}$ via scene decoder $Decoder_s$. This process updates the state to $\mathbf{S}_{t}$ and yields fused tokens $\mathbf{f}_{fused}$ which encapsulate depth-resolved spatial context:
\begin{align}
\label{eq:decoding_state}
[\mathbf{f}_{\text{fused}}, \mathbf{F}'_{\text{s}}], \mathbf{S}_t &= \operatorname{Decoder_s}([\mathbf{p}_h,\mathbf{F_{s}}], \mathbf{S}_{t-1}) \\
\label{eq:prediction_heads}
\mathcal{Y}_{\text{hand}} &= \left\{ \begin{aligned} &\operatorname{Head}_{\text{mano}}(\mathbf{f}_{\text{h}}) \\ &\operatorname{Head}_{\text{transl}}(\mathbf{f}_{\text{fused}}) \end{aligned} \right.
\end{align}

Based on the decoded features, we employ decoupled heads to maximize precision:
\begin{itemize}
    \item Global Positioning ($T$): The translation head operates on $\mathbf{f}_{fused}$. By utilizing the state-aware features, the model can resolve depth ambiguities to predict accurate absolute global translation.
    \item Local Articulation ($\theta, \beta$): The MANO head predicts pose and shape directly from the original expert tokens $\mathbf{f}_{h}$. This shortcut connection ensures that high-fidelity finger articulation cues are not diluted by the global fusion process.
\end{itemize}

\begin{table*}[t]
\caption{\textbf{Evaluation of local hand mesh reconstruction on DexYCB test set.} \\ The best and second-best results are highlighted in \textbf{bold} and \underline{underlined}, respectively.}
\label{tab:local_eval}
\centering
\footnotesize 
\renewcommand{\arraystretch}{0.9} 
\setlength{\tabcolsep}{0pt} 

% 4. 【关键】设置总宽度为 0.8\linewidth (或者 0.75\linewidth)，解决太宽的问题
\begin{tabular*}{0.9\linewidth}{@{\extracolsep{\fill}} clcccccc }
    \toprule
    \multirow{2}{*}{} & \multirow{2}{*}{\textbf{Methods}} & \multicolumn{2}{c}{\textbf{All}} & \multicolumn{2}{c}{\textbf{50\%-75\%}} & \multicolumn{2}{c}{\textbf{75\%-100\%}} \\
    \cmidrule(lr){3-4} \cmidrule(lr){5-6} \cmidrule(lr){7-8}
     & & \textbf{MPJPE}$\downarrow$ & \textbf{AUC}$\uparrow$ & \textbf{MPJPE}$\downarrow$ & \textbf{AUC}$\uparrow$ & \textbf{MPJPE}$\downarrow$ & \textbf{AUC}$\uparrow$ \\
    \midrule
    
    % --- Monocular Methods ---
    \multirow{5}{*}{\rotatebox[origin=c]{90}{monocular}}
     & Spurr \textit{et al.}~\cite{spurr2020weakly} & 6.83 & 86.4 & 8.00 & 84.0 & 10.65 & 78.8 \\
     & MeshGraphormer~\cite{lin2021mesh} & 6.41 & 87.2 & 7.22 & 85.6 & 7.76 & 84.5 \\
     & SemiHandObj~\cite{liu2021semi} & 6.33 & 87.4 & 7.17 & 85.7 & 8.96 & 82.1 \\
     & HandOccNet~\cite{park2022handoccnet} & 5.80 & 88.4 & 6.43 & 87.2 & 7.37 & 85.3 \\
     & WiLoR~\cite{potamias2024wilor} & 5.01 & \underline{90.0} & 5.42 & 89.2 & 5.68 & 88.7 \\
    \midrule
    
    % --- Temporal Methods ---
    \multirow{5}{*}{\rotatebox[origin=c]{90}{temporal}}
     & $S^2$HAND(V)~\cite{tu2023consistent} & 7.27 & 85.5 & 7.71 & 84.6 & 7.87 & 84.3 \\
     & VIBE~\cite{kocabas2020vibe} & 6.43 & 87.1 & 6.84 & 86.4 & 7.06 & 85.8 \\
     & TCMR~\cite{choi2021beyond} & 6.28 & 87.5 & 6.58 & 86.8 & 6.95 & 86.1 \\
     & Deformer~\cite{fu2023deformer} & 5.22 & 89.6 & 5.70 & 88.6 & 6.34 & 87.3 \\
     & HaWoR~\cite{zhang2025hawor} & \textbf{4.76} & \textbf{90.5} & \underline{5.03} & \underline{89.9} & \textbf{5.07} & \textbf{89.9} \\
    \midrule
    
    % --- Hand3R (Ours) ---
    \multirow{1}{*}{\rotatebox[origin=c]{90}{\textbf{ }}} 
     & \textbf{Hand3R (Ours)} & \underline{4.83} & \textbf{90.5} & \textbf{4.43} & \textbf{91.1} & \underline{5.52} & \underline{88.9} \\
    
    \bottomrule
\end{tabular*}
\end{table*}

\begin{table*}[t]
    \centering
    \caption{\textbf{Evaluation of global hand trajectory and reconstruction on HOI4D test set.} \\ Note that Hand3R is the only method that operates in an online, one-stage manner.}
    \label{tab:global_eval}
    
    % 使用 resizebox 确保表格适应页面宽度
    \resizebox{0.95\linewidth}{!}{
    \begin{tabular}{lcccccccc}
        \toprule
        % --- 表头第一行 ---
        \multirow{2}{*}{\textbf{Method}} & \multirow{2}{*}{\textbf{Type}} & \multirow{2}{*}{\textbf{Pipeline}} & \multirow{2}{*}{\textbf{C-MPJPE} $\downarrow$} & \multicolumn{2}{c}{\textbf{Short Video}} & \multicolumn{2}{c}{\textbf{Long Video}} \\
        
        % \cmidrule 用于画中间的横线，(lr) 表示左右留白，让分组更明显
        \cmidrule(lr){5-6} \cmidrule(lr){7-8}
        
        % --- 表头第二行 ---
         & & & & \textbf{WA-MPJPE} $\downarrow$ & \textbf{W-MPJPE} $\downarrow$ & \textbf{WA-MPJPE} $\downarrow$ & \textbf{W-MPJPE} $\downarrow$ \\
        \midrule
        
        % --- 对比方法 ---
        HaMeR-SLAM  & Offline & Multi-stage & 248.23 & 52.69 & 140.75 & 85.46 & 218.05 \\
        WiLoR-SLAM  & Offline & Multi-stage &    252.24    &   52.91   &  146.91    &   87.51  & 223.00  \\
        HaWoR & Offline & Multi-stage & 51.77 & 22.54 & 41.28 & 27.40 & 58.62 \\
        
        \midrule
        
        % --- 你的方法 ---
        Hand3R (Ours) & Online & One-stage & 42.6 & 38.04 & 86.87 & 56.71 & 125.81 \\
        \bottomrule
    \end{tabular}
    }
\end{table*}

\subsection{Training Strategy and Objectives}
\label{sec:training}

Training a unified 4D scene-hand reconstruction model requires balancing local hand mesh fidelity with global geometric consistency. We use a two-stage training strategy to progressively decouple these tasks.
\subsubsection{\textbf{Robust Pose Learning}}

In the first stage, we focus on establishing a robust hand pose prior. We freeze the scene-related branches and exclusively fine-tune the MANO Head. To maximize the diversity of hand articulations and improve generalization, we train on DexYCB\cite{chao2021dexycb}.
We employ relative geometric losses to supervise the hand pose regardless of its global position:
\begin{equation}
    \mathcal{L}_{\text{stage1}} = \lambda_{\text{joint}} ||\mathbf{J}_{\text{rel}} - \hat{\mathbf{J}}_{\text{rel}}||_2^2 + \lambda_{\text{vert}} ||\mathbf{V}_{\text{rel}} - \hat{\mathbf{V}}_{\text{rel}}||_2^2
\end{equation}
where $\mathbf{J}_{\text{rel}}$ and $\mathbf{V}_{\text{rel}}$ denote the root-relative coordinates of joints and vertices. 

\subsubsection{\textbf{Scene-Aware Global Tuning}}

In the second stage, we fine-tune the fused network and the translation head on the HOI4D~\cite{liu2022hoi4d} dataset to learn precise global positioning. The objective is formulated as:
\begin{equation}
    \mathcal{L}_{\text{stage2}} = {\mathcal{L}_{\text{trans}} + \mathcal{L}_{\text{abs}} + \mathcal{L}_{\text{2D}}} + \gamma {(\mathcal{L}_{\text{pts}} + \mathcal{L}_{\text{cam}})}
\end{equation}

Here, $\mathcal{L}_{\text{trans}}$ and $\mathcal{L}_{\text{abs}}$ minimize the $L_2$ errors of the global root translation and absolute 3D joint coordinates, respectively, while $\mathcal{L}_{\text{2D}}$ enforces pixel-level alignment via 2D keypoint reprojection. Crucially, we retain the scene objectives ($\mathcal{L}_{\text{pts}}$ and $\mathcal{L}_{\text{cam}}$) weighted by $\gamma$ to prevent catastrophic forgetting, ensuring the scene geometry remains consistent while the hand is accurately placed within it. 

Please refer to the supplementary materials A for a detailed analysis of the rationale behind our two-stage training strategy and dataset selection.

\section{Experiments}
\label{sec:exp}

\begin{figure*}[t] 
  \centering

  \includegraphics[width=1.02\linewidth, trim=0cm 2.1cm 0cm 2.2cm, clip]{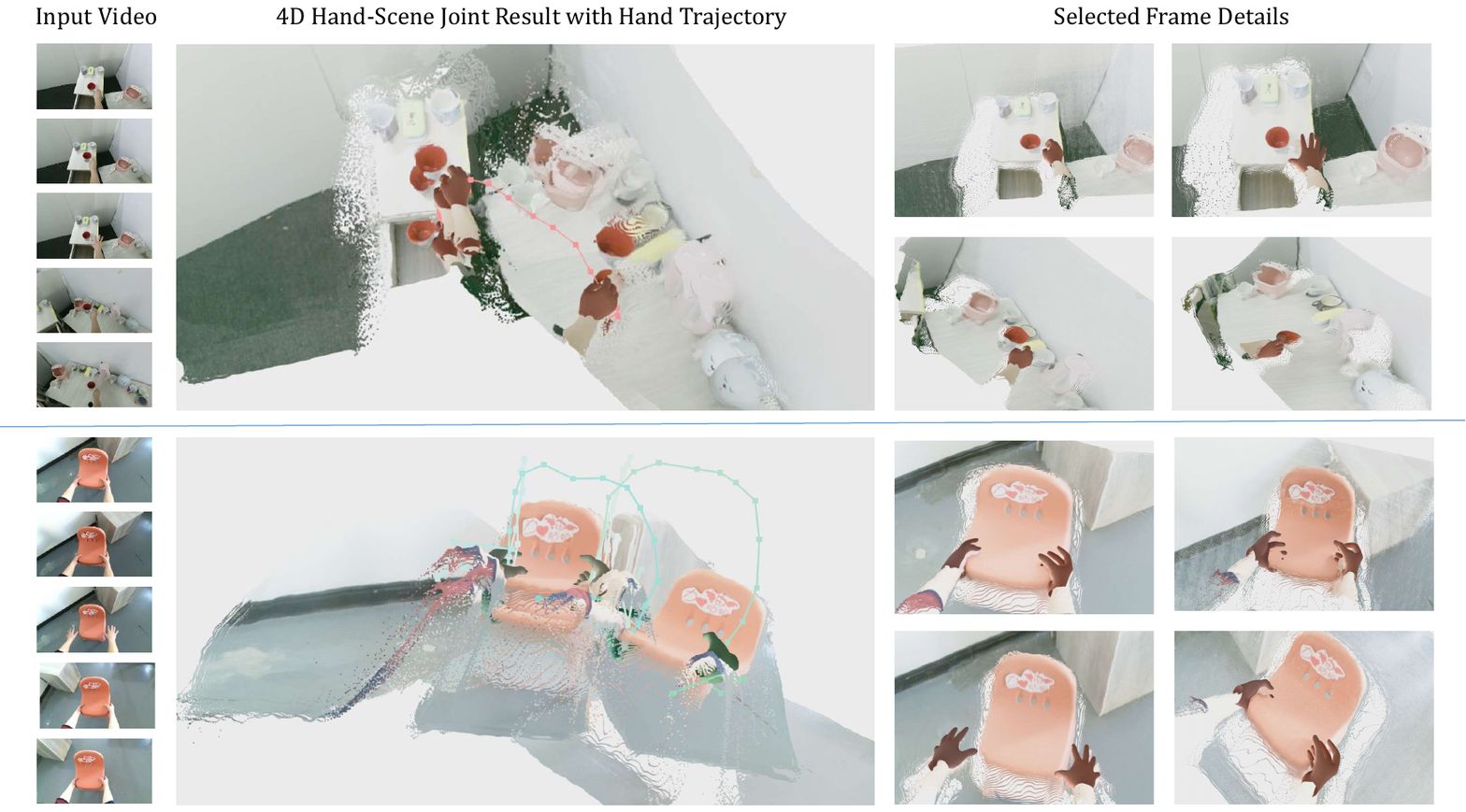}
  
\caption{\textbf{Qualitative visualization of hand-scene construction result to the unseen scenarios.} 
  We demonstrate Hand3R's robust performance on novel sequences that were not seen during training.
  \textbf{(Left)} Input video clips showing dynamic movement.
  \textbf{(Center)} The accumulated 4D reconstruction. We visualize the online-built dense scene geometry overlaid with global hand trajectory lines and sparsely sampled hand meshes. And our framework naturally supports multi-hand reconstruction, consistently tracking multi hands within the same global coordinate system.
  \textbf{(Right)} Detailed result at specific timestamps which highlight the fine-grained spatial accuracy.}
  \label{fig:qualitative_unseen}
\end{figure*}

We evaluate Hand3R on two primary tasks: local hand mesh recovery in camera coordinates (Sec.~\ref{sec:exp_local}) and online global hand reconstruction in world coordinates (Sec.~\ref{sec:exp_global}). Qualitative visualizations across unseen scenarios are provided in Sec.~\ref{sec:exp_qualitative}.

\subsection{Implementation Details}

For the hand expert, we adopt HaMeR~\cite{pavlakos2024reconstructing}. We utilize its pre-trained ViT backbone to extract high-fidelity tokens $\mathbf{f}_{h}$ that capture rich morphological details and robust geometric priors. 

For the scene foundation model, we build upon CUT3R~\cite{cut3r}. Specifically, we leverage its encoder to extract dense, metric-scale feature maps $\mathbf{F}_{s}$ from the full image, which serve as the spatial anchor for our prompting mechanism. Furthermore, we employ its $Decoder_s$ to integrate the visual prompts into the persistent memory state $\mathbf{S}_{t}$. This decoder fuses the current prompts with spatiotemporal history and current scene tokens, effectively resolving scale-depth ambiguity for precise global placement.

\subsection{Local Hand Mesh Recovery}
\label{sec:exp_local}

We first evaluate the fidelity of the reconstructed hand mesh in the local camera coordinate system. We follow the commonly used local hand mesh reconstruction benchmark, the DexYCB\cite{chao2021dexycb} dataset, which provides high-quality 3D hand annotations in controlled environments. 
Following standard protocols, we use Procrustes-Aligned MPJPE (PA-MPJPE) and the Area Under the Curve (AUC) of correctly localized keypoints as our primary metrics.
To assess the robustness of our model under severe occlusions, we further report performance on subsets categorized by occlusion ratios: $50\%-75\%$ and $75\%-100\%$.

As shown in Table~\ref{tab:local_eval}, Hand3R achieves competitive or superior performance compared to specialized local mesh regressors. Notably, despite being designed for global hand-scene reconstruction, our model maintains high local precision. We attribute this to our decoupled head design, where the local branch bypasses the global fusion module, preserving the fine-grained articulation features learned in robust pose learning.

\subsection{Global Hand Reconstruction}
\label{sec:exp_global}

We evaluate global hand reconstruction quality on HOI4D using three metrics: C-MPJPE measures absolute joint error in the camera frame; W-MPJPE measures global trajectory error after aligning the first frames of the sequence, and WA-MPJPE measures error after aligning the entire trajectory. We report these metrics on both short video (continuous 30 frames) and long Video (continuous 100 frames) sequences to assess tracking stability over time.

Since very few methods perform global hand reconstruction, we construct two strong baselines for comparison: HaMeR-SLAM and WiLoR-SLAM. Specifically, we combine the robust DROID-SLAM~\cite{teed2021droid} to obtain camera poses with SOTA hand estimators, HaMeR~\cite{pavlakos2024reconstructing} and WiLoR~\cite{potamias2024wilor}, to lift hands into the world frame.

As shown in Table~\ref{tab:global_eval}, Hand3R significantly outperforms the multi-stage offline baseline HaMeR-SLAM and WiLoR-SLAM across all metrics. When compared to the offline multi-stage method HaWoR, we observe a performance gap in global metrics. We attribute this disparity primarily to the structural distinction between our online, one-stage architecture and HaWoR's offline, multi-stage pipeline. 
Offline approaches inherently benefit from global bundle adjustment, which utilizes future frames to refine camera trajectories. In contrast, online methods are restricted to causal estimation, where accumulated drift is inevitable. 
This effect is particularly pronounced as sequence length increases from short videos to long videos, leading to an inherent precision gap in camera pose estimation that models cannot bridge.

Nevertheless, the strengths of Hand3R remain significant. Our method achieves the lowest C-MPJPE, demonstrating that it provides the most accurate absolute hand estimation in the camera frame, surpassing even the offline models. Furthermore, Hand3R comprehensively outperforms the HaMeR-SLAM and WiLoR-SLAM baselines, offering a robust high-performance solution for real-time applications.

\subsection{Qualitative Results}
\label{sec:exp_qualitative}

We provide qualitative visualizations in Fig.~\ref{fig:qualitative_unseen} to demonstrate the robustness of Hand3R on diverse, unseen in-the-wild scenarios. And we also provide additional animated visualizations in supplementary materials B.

These visualizations highlight two core strengths of our framework. 
First, they demonstrate the superior performance of our joint online hand-scene temporal reconstruction. As shown in the center view, Hand3R successfully integrates dynamic hand signals into the unified 4D scene representation. The seamless combination of the dense scene geometry and smooth hand trajectories indicates that our model effectively anchors hands within the persistent environment memory, achieving high fidelity in both temporal coherence and spatial alignment.
Second, the results exhibit robust generalization capabilities across diverse hand configurations. Our framework naturally supports both multi-hand and left-hand scenarios, consistently tracking and reconstructing these complex cases within the same metric space with high accuracy.

\section{Conclusion}
\label{sec:conclusion}

We presented Hand3R, the first online, end-to-end framework for joint 4D hand-scene reconstruction. 
By leveraging the scene-aware visual prompting mechanism, our model effectively synergizes high-fidelity hand priors with persistent metric-scale scene memory. 
Experiments demonstrate that Hand3R successfully performs joint hand-scene reconstruction while maintaining competitive performance against specialized baselines.
We believe Hand3R establishes a strong foundation for real-time Embodied AI and AR/VR applications where simultaneous hand-scene understanding is paramount.

\bibliographystyle{IEEEbib}
\bibliography{icme2026references}

\end{document}